\documentclass[conference]{IEEEtran}
\IEEEoverridecommandlockouts

\usepackage{cite}
\usepackage{amsmath,amssymb,amsfonts}
\usepackage{algorithmic}
\usepackage{graphicx}
\usepackage{textcomp}
\usepackage{xcolor}
\usepackage{booktabs}
\usepackage{multirow}
\usepackage{url}

\usepackage{enumitem}
\usepackage[draft,bookmarks=false]{hyperref}

\setlist[itemize]{topsep=1.5pt, itemsep=1pt, parsep=0pt, partopsep=0pt}
\setlist[enumerate]{topsep=1.5pt, itemsep=1pt, parsep=0pt, partopsep=0pt}

\AtBeginDocument{%
  \setlength{\abovedisplayskip}{4pt plus 1pt minus 1pt}%
  \setlength{\belowdisplayskip}{4pt plus 1pt minus 1pt}%
  \setlength{\abovedisplayshortskip}{2pt plus 1pt minus 1pt}%
  \setlength{\belowdisplayshortskip}{2pt plus 1pt minus 1pt}%
}

\usepackage{titlesec}
\titlespacing*{\section}{0pt}{4.5pt plus 1pt minus 1pt}{2pt plus 0.5pt minus 0.5pt}
\titlespacing*{\subsection}{0pt}{3.5pt plus 1pt minus 1pt}{1.5pt plus 0.5pt minus 0.5pt}

\begin{document}

\title{\vspace*{-0.5em}When Do LLMs Help with Feature Selection?\\
A Systematic Empirical Study\vspace{-0.3em}}
\author{%
\makebox[\textwidth][c]{%
\begin{minipage}[t]{0.33\textwidth}
\centering
\textbf{Fateme Memar}\\
\textit{Department of Electrical Engineering\\
and Computer Science}\\
\textit{University of Kansas}\\
Lawrence, KS 66045, USA\\
amemar@ku.edu
\end{minipage}
\hfill
\begin{minipage}[t]{0.33\textwidth}
\centering
\textbf{Dongjie Wang}\\
\textit{School of Information Science\\
and Technology}\\
\textit{Northeast Normal University}\\
China\\
wangdongjie100@nenu.edu.cn
\end{minipage}
\hfill
\begin{minipage}[t]{0.33\textwidth}
\centering
\textbf{Arvin Agah}\\
\textit{Department of Electrical Engineering\\
and Computer Science}\\
\textit{University of Kansas}\\
Lawrence, KS 66045, USA\\
agah@ku.edu
\end{minipage}%
}\vspace{-0.6em}%
}
\maketitle

\begin{abstract}
Large language models (LLMs) can leverage semantic feature names alongside summary statistics for tabular feature selection, yet it remains unclear when such information improves subset selection relative to established statistical methods. We investigate this question within an 18-dataset framework comprising 10 datasets for Group Relative Policy Optimization (GRPO) training, 2 for checkpoint validation, and 6 held-out evaluation datasets. We compare five classical baselines, frozen Qwen2.5-7B and Llama-3.3-70B selectors, and two GRPO-adapted Qwen2.5-7B models, reporting classification ROC-AUC and regression $R^2$ separately and using dataset-wise ranks for cross-task comparison. Across five seeded evaluation partitions, classical selectors generally achieve higher predictive performance than the tested 7B LLM/RL selectors, including under a pre-specified ${\sim}33\%$ feature budget with strict train-only classical fitting, while the 7B selectors retain compact subsets. A cardinality-matched semantic sensitivity experiment shows that, for Zero-shot Qwen-7B, informative feature names improve regression performance over anonymized and shuffled alternatives ($+0.1964$ and $+0.1606$ mean $R^2$), whereas aggregate classification differences are smaller ($+0.0159$ and $+0.0084$ ROC-AUC) and dataset-dependent. These results characterize when semantic information benefits LLM-based feature selection and show that classical selectors remain strong baselines for predictive tabular feature selection.
\end{abstract}


\section{Introduction}
\label{sec:intro}

Feature selection---identifying a compact, informative subset of features before training a downstream predictor---is a foundational preprocessing decision in tabular machine learning~\cite{guyon2003introduction,saeys2007review}. Irrelevant or redundant variables increase computational cost, obscure interpretability, and degrade generalization. Classical selectors such as minimum Redundancy Maximum Relevance (mRMR)~\cite{ding2005mrmr}, LASSO~\cite{tibshirani1996regression}, Boruta~\cite{kursa2010boruta}, and tree-based importance~\cite{chen2016xgboost} estimate feature utility from statistical relationships between inputs and targets, without explicitly exploiting real-world meaning in feature names.

Large language models (LLMs) introduce a different information source for feature selection~\cite{brown2020gpt3,vaswani2017attention}. Models such as Qwen2.5-7B~\cite{qwen2025qwen25} and Llama-3.3-70B~\cite{grattafiori2024llama3} combine semantic priors in feature names with coarse distributional summaries to propose compact subsets~\cite{choi2022lmpriors,jeong2024llmselect}. Although recent work has explored LLM-based feature engineering and tabular reasoning~\cite{hollmann2023caafe,han2024featllm,hegselmann2023tabllm,zhang2024tablellama}, an important question remains: \emph{what information do LLM feature selectors exploit, when do semantic priors help or mislead, and how do their sparsity--performance trade-offs compare with classical selectors under controlled information and feature-budget conditions?}

\begin{figure*}[t]
  \centering
  \includegraphics[
    width=\textwidth,
    trim=36 30 36 20,
    clip
  ]{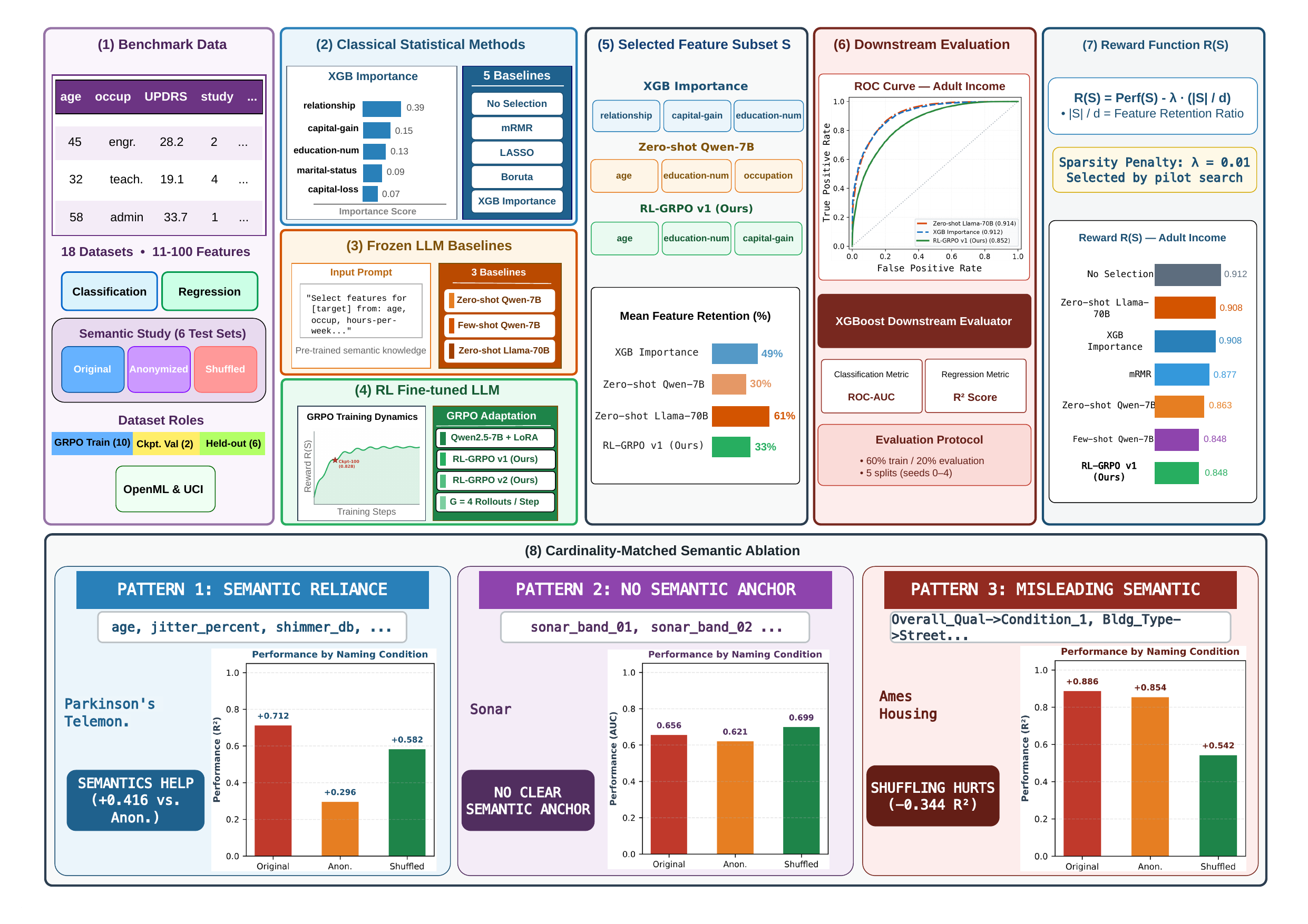}
  \caption{Overview of the experimental framework. The 18 datasets are partitioned into 10 for GRPO training, 2 for checkpoint validation, and 6 held-out evaluation datasets. Classical, frozen-LLM, and GRPO-adapted selectors are compared using task-separated ROC-AUC and $R^2$, with additional cardinality-matched semantic and train-only matched-budget controls.}
  \label{fig:framework}
\end{figure*}

We organize the study around three research questions:
\begin{itemize}
  \item \textbf{RQ1 (Performance and sparsity):} How do frozen-LLM, GRPO-adapted, and classical selectors compare 
  across held-out tabular datasets in predictive performance and feature retention?
  \item \textbf{RQ2 (Semantic mechanism):} How much do meaningful feature names contribute when subset cardinality is controlled and dataset and target identities are masked?
  \item \textbf{RQ3 (Robustness and controlled baselines):} Are the comparisons stable across five seeded evaluation partitions, task-separated evaluation, and train-only classical selectors under an approximately matched feature budget?
\end{itemize}

To address these questions, we combine a held-out benchmark with two controlled experiments. A cardinality-matched semantic diagnostic compares \emph{Original}, \emph{Anonymized}, and \emph{Shuffled} feature identities while masking dataset and target identities, and a train-only classical top-$k$ control evaluates classical selectors under a one-third feature budget. Classification ROC-AUC and regression $R^2$ are analyzed separately, using dataset-wise ranks for cross-task summaries and five seeded evaluation partitions to quantify variability.

\textbf{Key findings.} First, classical selectors generally provide stronger predictive performance than the tested LLM/RL selectors, including under the matched-budget, train-only control, while the 7B selectors produce compact subsets. Second, after controlling subset cardinality, meaningful feature names substantially improve Zero-shot Qwen-7B selection on the regression tasks, whereas classification effects are smaller and dataset-dependent. Third, GRPO adaptation changes selection behavior but does not yield consistent predictive superiority over either frozen LLM or classical alternatives.

\textbf{Contributions.} Rather than positioning LLMs as replacements for classical feature selectors, we characterize how semantic information, sparsity constraints, and RL adaptation interact in tabular feature selection:
\begin{enumerate}
  \item \textbf{Controlled feature-selection benchmark:} We compare classical, frozen-LLM, and GRPO-adapted selectors in an 18-dataset training/validation/held-out framework.
  \item \textbf{Cardinality-matched semantic diagnostic:} We introduce an exact-$k$ three-condition experiment separating feature-name semantics from subset-size confounding.
 \item \textbf{Robustness and matched-budget controls:} We use five-partition uncertainty, task-separated rank evaluation, and train-only classical top-$k$ controls.
\end{enumerate}

\section{Related Work}
\label{sec:related}

\subsection{Classical Feature Selection}

Feature-selection methods are commonly categorized as filter, wrapper, and embedded approaches~\cite{guyon2003introduction,saeys2007review,li2024feature,kohavi1997wrappers}. Representative methods include mRMR~\cite{ding2005mrmr}, which balances target relevance and inter-feature redundancy; Boruta~\cite{kursa2010boruta}, an all-relevant random-forest wrapper; LASSO~\cite{tibshirani1996regression}, which induces sparsity through $\ell_1$ regularization; and XGBoost importance~\cite{chen2016xgboost}, which ranks features by tree-based gain. Unlike post-hoc attribution methods such as SHAP~\cite{lundberg2017shap} and LIME~\cite{ribeiro2016lime}, our study focuses on pre-training subset selection and uses these four selectors alongside a \emph{No Selection} control.

\subsection{Large Language Models for Tabular and Feature Tasks}

LLMs are increasingly applied to tabular prediction and feature engineering. TabLLM~\cite{hegselmann2023tabllm}, LIFT~\cite{dinh2022lift}, and TableLlama~\cite{zhang2024tablellama} adapt LLMs to tabular tasks, while LMPriors~\cite{choi2022lmpriors}, CAAFE~\cite{hollmann2023caafe}, FeatLLM~\cite{han2024featllm}, and LLM-Select~\cite{jeong2024llmselect} use semantic metadata for feature construction or selection. However, the effect of feature-name semantics remains difficult to separate from subset size and evaluation protocol. We address this gap via cardinality-matched semantic interventions and controlled comparisons with classical and GRPO-adapted selectors.

\subsection{Reinforcement Learning for LLM Fine-Tuning}

Reinforcement learning is widely used to align language models with task objectives~\cite{ouyang2022instructgpt}. PPO~\cite{schulman2017ppo}, DPO~\cite{rafailov2023dpo}, and KTO~\cite{ethayarajh2024kto} provide optimization paradigms, while GRPO~\cite{shao2024deepseekmath,deepseek2025r1} estimates relative advantages from groups of sampled completions without a separate value critic. We combine GRPO with LoRA~\cite{hu2021lora} using TRL~\cite{huggingface2023trl} to optimize a downstream performance--sparsity objective for feature selection.

\subsection{Tabular Data Learning}

Tree-based ensembles remain highly competitive on heterogeneous tabular data~\cite{shwartzziv2022tabular,grinsztajn2022treebased,mcelfresh2023neural}. Neural alternatives include TabNet~\cite{arik2021tabnet}, TabTransformer~\cite{huang2020tabtransformer}, FT-Transformer~\cite{gorishniy2021revisiting}, SAINT~\cite{somepalli2022saint}, and the TabPFN foundation-model family~\cite{muller2025tabpfnv2}. Given the continued strength of boosted trees, we use XGBoost~\cite{chen2016xgboost} as a fixed downstream evaluator across OpenML~\cite{vanschoren2014openml} and UCI~\cite{dua2017uci} datasets.

\section{Problem Formulation}
\label{sec:problem}

\subsection{Feature Selection as Subset Search}

Let $\mathcal{D}=\{(\mathbf{x}_i,y_i)\}_{i=1}^{n}$ contain $n$ observations and $d$ candidate features $\mathcal{F}=\{f_1,\ldots,f_d\}$. A selector produces $S\subseteq\mathcal{F}$, evaluated by predictive performance and feature retention $|S|/d$~\cite{guyon2003introduction}. For classification and regression tasks,
\begin{equation}
\mathrm{Perf}(S)=
\begin{cases}
\mathrm{AUC}(S), & \text{classification},\\
R^2(S), & \text{regression}.
\end{cases}
\label{eq:performance}
\end{equation}
ROC-AUC and $R^2$ are analyzed separately, with dataset-wise ranks used for cross-task summaries.

\subsection{Performance--Sparsity Reward}

We define
\begin{equation}
R(S)=\mathrm{Perf}(S)-\lambda\frac{|S|}{d},
\label{eq:reward}
\end{equation}
where $\lambda$ controls the sparsity penalty. We use $\lambda=0.01$, selected by pilot search, and evaluate subsets with a fixed XGBoost downstream predictor~\cite{chen2016xgboost}. This reward guides GRPO training, while ROC-AUC and $R^2$ remain the primary metrics.

\subsection{Feature-Selection Paradigms}

We compare three mechanisms for constructing $S$:

\textbf{1) Classical Statistical Selection.}
mRMR~\cite{ding2005mrmr}, LASSO~\cite{tibshirani1996regression}, Boruta~\cite{kursa2010boruta}, and XGBoost importance~\cite{chen2016xgboost} infer feature utility from tabular relationships; \emph{No Selection} retains all features.

\textbf{2) Frozen LLM Selection.}
Zero-shot and Few-shot Qwen-7B~\cite{qwen2025qwen25} and Zero-shot Llama-70B~\cite{grattafiori2024llama3} select features from feature names and summary statistics without fine-tuning.

\textbf{3) GRPO-Adapted LLM Selection.}
Qwen2.5-7B is adapted with GRPO~\cite{shao2024deepseekmath,deepseek2025r1} and LoRA~\cite{hu2021lora} using Eq.~\eqref{eq:reward}.

\subsection{Cardinality-Matched Semantic Diagnostic}

To isolate feature-name semantics from subset size, we evaluate Zero-shot Qwen-7B under \emph{Original}, \emph{Anonymized}, and \emph{Shuffled} features with dataset and target identities masked. For dataset $d$ and completion $c$, cardinality is matched:
\begin{equation}
|S^{\mathrm{orig}}_{d,c}|=
|S^{\mathrm{anon}}_{d,c}|=
|S^{\mathrm{shuf}}_{d,c}|.
\label{eq:matched_k}
\end{equation}
Original--Anonymized differences capture informative feature semantics, while Original--Shuffled differences capture sensitivity to misleading semantics at fixed cardinality.
\section{Method}
\label{sec:method}

\subsection{LLM Feature-Selection Pipeline}

\textbf{Prompt construction.}
For each dataset, \texttt{PromptBuilder} creates a prompt containing dataset/task metadata and per-feature summaries. Numeric features include mean, standard deviation, range, and missing rate; categorical features include the top three category frequencies and missing rate. No individual rows, instance-level targets, or external descriptions are provided. Summary statistics are computed at the dataset level before downstream partitioning. Few-shot Qwen-7B prepends two exemplars drawn only from the 10 training datasets, using the top $\lfloor d/2 \rfloor$ XGBoost-ranked features.

\textbf{Output parsing.}
Using \texttt{tag\_first=True}, all LLM/RL selectors return
\begin{center}
\small\texttt{SELECTED\_FEATURES: ["feature\_a", ...]}
\end{center}
\texttt{OutputParser} extracts the final tagged list, validates tokens against the feature set $V$, removes duplicates, and discards hallucinated names. Missing, invalid, or empty outputs trigger a keep-all fallback ($S=V$) and are flagged with \texttt{parse\_ok=False}.

\subsection{GRPO Adaptation}

We use Qwen2.5-7B-Instruct~\cite{qwen2025qwen25} as the RL base model and evaluate frozen Qwen-7B and 4-bit Llama-3.3-70B-Instruct~\cite{grattafiori2024llama3} baselines. LoRA~\cite{hu2021lora} is applied to \texttt{q\_proj}, \texttt{k\_proj}, \texttt{v\_proj}, and \texttt{o\_proj} with $r=16$, $\alpha=32$, dropout $0.05$, and no bias.

GRPO~\cite{shao2024deepseekmath,deepseek2025r1} is implemented with TRL~\cite{huggingface2023trl}. For each prompt, the policy generates $G=4$ completions, each parsed into $S_g$ and scored by Eq.~\eqref{eq:reward}. Rewards are averaged across five seeded partitions of the current training dataset and normalized within the group to form relative advantages. Updates use KL regularization with $\beta=0.1$ against the frozen reference model. Repeated subset evaluations are cached to avoid redundant XGBoost scoring.

\subsection{Training and Checkpoint Selection}

The 18 datasets are split into 10 GRPO-training, 2 validation, and 6 held-out evaluation datasets; held-out datasets are excluded from GRPO training, few-shot exemplars, and checkpoint selection. The sparsity weight $\lambda=0.01$ is selected from $\{0.01,0.02,0.03,0.05,0.10,0.15,0.20\}$ on \textit{heart\_disease} and fixed thereafter.

Both GRPO variants use Qwen2.5-7B-Instruct with batch size 1, learning rate $10^{-6}$, maximum gradient norm $0.1$, gradient checkpointing, and maximum completion length $1024$.

\begin{itemize}
  \item \textbf{RL-GRPO v1:} Trained for 500 steps with temperature $0.7$ and checkpoints every 100 steps. Checkpoint-100 attains the highest mean validation reward ($0.8282$) on \textit{mice\_protein} and \textit{communities\_crime} and is selected without consulting held-out datasets.
  \item \textbf{RL-GRPO v2:} Trained separately for 1,500 steps from the base model with newly initialized LoRA adapters and evaluated at the terminal checkpoint without validation-based checkpoint selection. Thus, v1 and v2 represent distinct GRPO configurations rather than an isolated training-duration comparison.
\end{itemize}

\begin{figure}[t]
  \centering
  \includegraphics[width=\columnwidth]{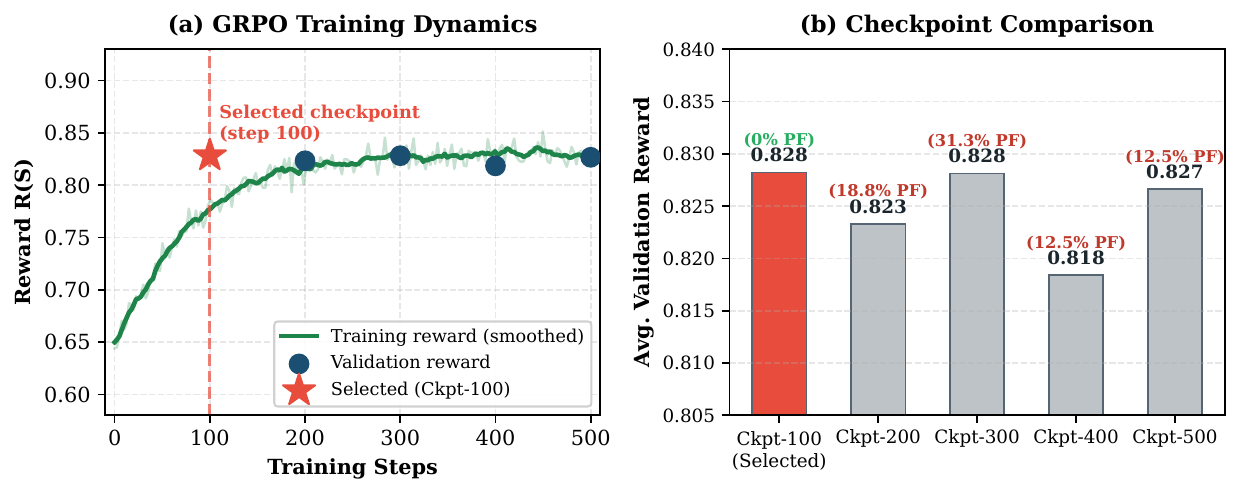}
  \caption{RL-GRPO v1 training and validation rewards across checkpoints 100--500; Checkpoint-100 achieves the highest mean validation reward ($0.8282$).}
  \label{fig:training}
\end{figure}

\subsection{Inference and Reproducibility}

Each LLM/RL selector generates $K=4$ stochastic completions per dataset with temperature $0.7$ and \texttt{max\_new\_tokens=512}. The four completion scores are averaged within each of five seeded evaluation partitions (seeds $0$--$4$) to yield one observation per dataset--method pair and avoid pseudoreplication.
Selection consistency is measured by mean pairwise Jaccard similarity over valid completions. Experiments use PyTorch, HuggingFace \texttt{transformers}, \texttt{peft}, \texttt{trl} v0.29.1, \texttt{scikit-learn}~\cite{pedregosa2011scikit}, and \texttt{xgboost}~\cite{chen2016xgboost}.


\section{Experimental Setup}
\label{sec:setup}

\subsection{Datasets and Dataset Roles}
\label{sec:setup_datasets}

We use 18 OpenML~\cite{vanschoren2014openml} and UCI~\cite{dua2017uci} datasets: 10 for GRPO training (\textit{heart\_disease}, \textit{german\_credit}, \textit{wine\_quality}, \textit{bank\_marketing}, \textit{breast\_cancer}, \textit{dermatology}, \textit{ionosphere}, \textit{spambase}, \textit{optical\_digits}, \textit{geo\_music}), 2 for checkpoint validation (\textit{mice\_protein}, $d=77$; \textit{communities\_crime}, $d=100$), and 6 for held-out evaluation (\textit{adult\_income}, $d=13$; \textit{parkinsons\_telemonitoring}, $d=19$; \textit{student\_performance}, $d=30$; \textit{student\_dropout}, $d=36$; \textit{sonar}, $d=60$; \textit{ames\_housing}, $d=80$), spanning classification and regression ($d=11$--$100$). Held-out datasets are excluded from GRPO training, few-shot exemplars, reward updates, and checkpoint selection.

\subsection{Compared Methods}
\label{sec:setup_methods}

We compare ten configurations: \textit{No Selection}; mRMR~\cite{ding2005mrmr} and XGB Importance~\cite{chen2016xgboost} ($\lfloor d/2 \rfloor$ features), LASSO~\cite{tibshirani1996regression}, and Boruta~\cite{kursa2010boruta}; frozen Zero-shot and Few-shot Qwen-7B~\cite{qwen2025qwen25} and Zero-shot Llama-70B~\cite{grattafiori2024llama3}; and RL-GRPO v1/v2.

In the primary benchmark, classical selectors are fitted on the full preprocessed dataset before downstream partitioning, whereas LLM selectors receive dataset-level summaries rather than row-level instances and labels. We therefore include a strict train-only matched-budget classical control (Section~\ref{sec:setup_trainonly}) to address this information-regime asymmetry.

\subsection{Evaluation Protocol and Metrics}
\label{sec:setup_metrics}

All methods are evaluated on the six held-out datasets across five seeded evaluation partitions (seeds $0$--$4$); seed $42$ is retained only as a reproduction reference and excluded from the $N=5$ sample. Each partition uses 60\% training, 20\% evaluation, and 20\% unused data, stratified for classification. A fixed XGBoost~\cite{chen2016xgboost} model (100 estimators, maximum depth 4, learning rate $0.1$, row/column subsample $0.8$, seed $s$) is trained on the 60\% training split using the selected subset $S$ and evaluated on the corresponding 20\% split.

For stochastic LLM/RL selectors, the $K=4$ completion scores are averaged within each partition first, yielding $N=5$ partition-level observations per dataset--method pair. We report classification ROC-AUC and regression $R^2$ separately, feature retention $|S|/d$, reward $R(S)$, and dataset-wise performance ranks for cross-task summaries. Across partitions, we report mean$\pm$SD and descriptive $t$-based 95\% confidence intervals; LLM/RL reliability is summarized by parse-failure rate and mean pairwise Jaccard similarity over valid completions.

\subsection{Cardinality-Matched Semantic Diagnostic}
\label{sec:setup_semantic}

Zero-shot Qwen-7B is evaluated under \emph{Original}, \emph{Anonymized}, and \emph{Shuffled} feature-name conditions across all six held-out datasets, with $K=4$ completions per condition and the same five seeded evaluation partitions. Dataset and target names are masked in all conditions. For each dataset $d$ and completion $c$, all three conditions use the exact cardinality $k_{d,c}$ of the corresponding Original completion, enforcing $\Delta\mathrm{Retention}=0\%$. Shuffled names are assigned by a strict derangement with zero fixed points and mapped back to their true columns before evaluation. Prompt summary statistics remain unchanged across conditions. Selections are ordered by importance and truncated to $k_{d,c}$ when necessary; outputs with fewer than $k_{d,c}$ valid features are retried up to three times without random padding. No parse or exact-$k$ failures occurred across the 72 generations.

\subsection{Train-Only Matched-Budget Classical Control}
\label{sec:setup_trainonly}

To control both information regime and subset size, we evaluate train-only classical selectors at a pre-specified budget matching RL-GRPO v1's 33.37\% mean retention:
\begin{equation}
k_d=\max\!\big(1,\operatorname{round}(0.3337d)\big).
\label{eq:matched_budget}
\end{equation}
This yields 32.68\% mean realized retention across the six held-out datasets. Within each seed-specific partition (after initial dataset-level missing-value filtering and categorical encoding), mRMR Top-$k$, XGB Importance Top-$k$, and LASSO Top-$k$ are fitted strictly on the 60\% training split with no access to evaluation rows or labels; LASSO standardization is also fit train-only. LASSO ranks features by standardized coefficient magnitude. Boruta is omitted because its all-relevant procedure does not provide a strict top-$k$ ordering.

\section{Results}
\label{sec:results}

\subsection{Overall Comparison}
\label{sec:results:main}

Table~\ref{tab:main} reports performance across five seeded evaluation partitions for all 10 methods on the six held-out datasets. ROC-AUC and $R^2$ are reported separately, with dataset-wise ranks averaged across the six datasets for cross-task comparison and feature retention used to characterize sparsity.

\begin{table*}[t]
\centering
\caption{Predictive performance across the 6 held-out datasets over $N=5$ seeded evaluation partitions (mean $\pm$ SD). Classification reports ROC-AUC and regression reports $R^2$; summary columns give task-specific means, 6-dataset mean rank ($1=\text{best}$), and mean feature retention. Best predictive-performance and rank values are \textbf{bold}.}
\label{tab:main}
\setlength{\tabcolsep}{2.8pt}
\renewcommand{\arraystretch}{1.15}
\resizebox{\textwidth}{!}{%
\begin{tabular}{ll ccc ccc ccc c}
\toprule
& & \multicolumn{3}{c}{\textbf{Classification (ROC-AUC)}} & \multicolumn{3}{c}{\textbf{Regression ($R^2$)}} & \multicolumn{3}{c}{\textbf{Task-Separated Summary}} & \\
\cmidrule(lr){3-5} \cmidrule(lr){6-8} \cmidrule(lr){9-11}
\textbf{Method} & \textbf{Category} & \textbf{Adult} & \textbf{Sonar} & \textbf{Dropout} & \textbf{Ames} & \textbf{Park.} & \textbf{Perform.} & \textbf{Clf AUC} & \textbf{Reg $R^2$} & \textbf{Rank}$\downarrow$ & \textbf{Feat\%} \\
\midrule
No Selection & Classical & $\mathbf{0.922{\pm}.003}$ & $0.912{\pm}.042$ & $\mathbf{0.894{\pm}.008}$ & $0.915{\pm}.009$ & $0.884{\pm}.005$ & $\mathbf{0.305{\pm}.056}$ & $0.909$ & $\mathbf{0.701}$ & $\mathbf{2.67}$ & $100\%$ \\
mRMR~\cite{ding2005mrmr} & Classical & $0.882{\pm}.004$ & $0.899{\pm}.067$ & $0.886{\pm}.009$ & $0.911{\pm}.010$ & $\mathbf{0.896{\pm}.006}$ & $0.269{\pm}.043$ & $0.889$ & $0.692$ & $3.67$ & $49\%$ \\
LASSO~\cite{tibshirani1996regression} & Classical & $\mathbf{0.922{\pm}.003}$ & $\mathbf{0.914{\pm}.036}$ & $0.893{\pm}.009$ & $0.911{\pm}.008$ & $0.671{\pm}.013$ & $0.278{\pm}.054$ & $\mathbf{0.910}$ & $0.620$ & $3.25$ & $67\%$ \\
Boruta~\cite{kursa2010boruta} & Classical & $0.903{\pm}.003$ & $0.913{\pm}.053$ & $0.881{\pm}.009$ & $0.907{\pm}.010$ & $0.884{\pm}.005$ & $0.195{\pm}.057$ & $0.899$ & $0.662$ & $5.08$ & $38\%$ \\
XGB Imp.~\cite{chen2016xgboost} & Classical & $0.912{\pm}.002$ & $0.887{\pm}.026$ & $0.891{\pm}.008$ & $\mathbf{0.918{\pm}.006}$ & $0.890{\pm}.015$ & $0.258{\pm}.105$ & $0.897$ & $0.689$ & $3.50$ & $49\%$ \\
\midrule
Zero-shot Qwen-7B~\cite{qwen2025qwen25} & Frozen LLM & $0.867{\pm}.005$ & $0.778{\pm}.046$ & $0.812{\pm}.009$ & $0.863{\pm}.021$ & $0.889{\pm}.009$ & $-0.039{\pm}.044$ & $0.819$ & $0.571$ & $7.83$ & $30\%$ \\
Few-shot Qwen-7B~\cite{qwen2025qwen25} & Frozen LLM & $0.852{\pm}.006$ & $0.803{\pm}.045$ & $0.813{\pm}.007$ & $0.865{\pm}.018$ & $0.891{\pm}.013$ & $-0.038{\pm}.043$ & $0.823$ & $0.573$ & $7.00$ & $32\%$ \\
Zero-shot Llama-70B~\cite{grattafiori2024llama3} & Frozen LLM & $0.914{\pm}.003$ & $0.864{\pm}.059$ & $0.856{\pm}.009$ & $0.903{\pm}.011$ & $0.886{\pm}.005$ & $0.258{\pm}.064$ & $0.878$ & $0.682$ & $5.17$ & $61\%$ \\
\midrule
\textbf{RL-GRPO v1 (Ours)}~\cite{shao2024deepseekmath} & \textbf{RL Fine-Tuned} & $0.852{\pm}.005$ & $0.726{\pm}.057$ & $0.819{\pm}.009$ & $0.873{\pm}.015$ & $0.888{\pm}.012$ & $+0.027{\pm}.027$ & $0.799$ & $0.596$ & $7.67$ & $33\%$ \\
\textbf{RL-GRPO v2 (Ours)}~\cite{shao2024deepseekmath} & \textbf{RL Fine-Tuned} & $0.856{\pm}.005$ & $0.794{\pm}.051$ & $0.801{\pm}.008$ & $0.860{\pm}.013$ & $0.883{\pm}.007$ & $-0.043{\pm}.042$ & $0.817$ & $0.567$ & $9.17$ & $32\%$ \\
\bottomrule
\end{tabular}%
}
\end{table*}

\begin{figure*}[t]
\centering
\includegraphics[width=\textwidth]{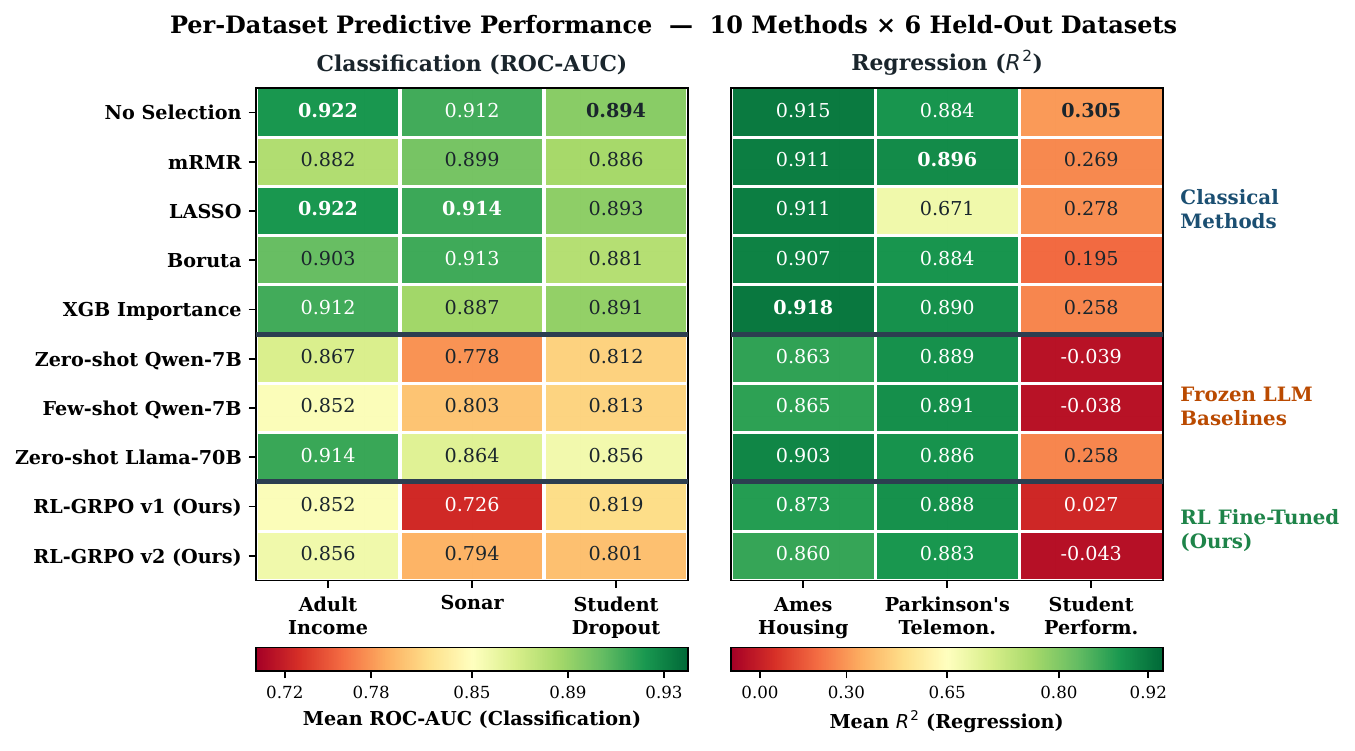}
\caption{Per-dataset mean predictive performance across the 10 methods over $N=5$ seeded evaluation partitions.}
\label{fig:heatmap}
\end{figure*}

\textbf{1) RL Fine-Tuning vs.\ Base Model.}
RL-GRPO v1 improves mean regression $R^2$ over frozen Qwen-7B ($0.596$ vs.\ $0.571$), mitigating its negative $R^2$ on \textit{student\_performance} ($+0.027$ vs.\ $-0.039$), improving \textit{ames\_housing} ($0.873$ vs.\ $0.863$), and remaining similar on \textit{parkinsons\_telemonitoring} ($0.888$ vs.\ $0.889$). On classification, v1 improves on \textit{student\_dropout} ($0.819$ vs.\ $0.812$) but decreases on \textit{adult\_income} ($0.852$ vs.\ $0.867$) and \textit{sonar} ($0.726$ vs.\ $0.778$), lowering ROC-AUC from $0.819$ to $0.799$. Thus, GRPO adaptation produces a task-dependent trade-off rather than uniform improvement.

\textbf{2) RL Fine-Tuning vs.\ Large Frozen LLM.}
Zero-shot Llama-70B achieves stronger aggregate performance than RL-GRPO v1, with mean classification ROC-AUC $0.878$ vs.\ $0.799$, regression $R^2$ $0.682$ vs.\ $0.596$, and rank $5.17$ vs.\ $7.67$. The exception is \textit{parkinsons\_telemonitoring}, where v1 slightly exceeds Llama-70B ($0.888$ vs.\ $0.886$). This advantage is accompanied by denser selection ($60.8\%$ vs.\ $33.4\%$ retention).

\textbf{3) Five-Partition Uncertainty and Parse Reliability.}
With only $N=5$ partitions, even a unanimous $5$--$0$ paired difference yields a minimum two-sided exact Wilcoxon $p=0.0625$ before multiplicity correction; comparisons are therefore interpreted descriptively rather than as evidence of statistical significance. Parse failures were $0/24$ for Zero-shot Qwen-7B and $1/24$ ($4.2\%$) for each remaining LLM/RL method.

Figure~\ref{fig:heatmap} shows the same per-dataset pattern summarized in Table~\ref{tab:main}: classical selectors generally rank more strongly, whereas the 7B LLM selectors retain substantially fewer features. We therefore next examine the performance--sparsity trade-off and a matched-budget classical control.

\subsection{Performance--Sparsity Trade-off}
\label{sec:results:pareto}

\begin{figure}[t]
\centering
\includegraphics[width=\columnwidth]{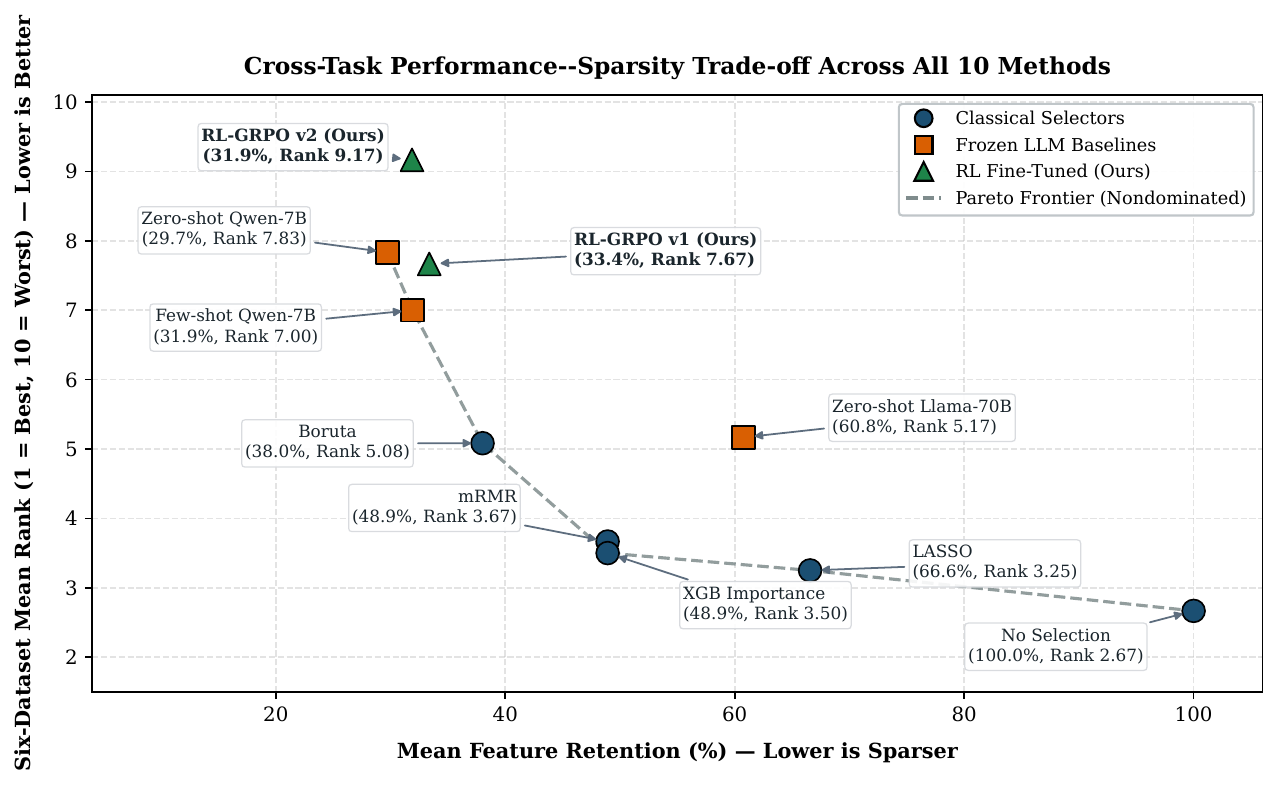}
\caption{Performance--sparsity trade-off using six-dataset mean rank and mean feature retention; lower values are preferable on both axes.}
\label{fig:pareto}
\end{figure}

Figure~\ref{fig:pareto} shows a trade-off rather than a uniformly dominant selector. Classical methods achieve stronger predictive ranks with higher retention, whereas the 7B LLM selectors retain $29.7\%$--$33.4\%$ of features but rank more weakly overall. The Pareto frontier includes Zero-shot Qwen-7B, Few-shot Qwen-7B, Boruta, XGB Importance, LASSO, and No Selection. Among 7B methods, Few-shot Qwen-7B achieves rank $7.00$ at $31.9\%$ retention, compared with $7.67$ at $33.4\%$ for RL-GRPO v1 and $9.17$ at $31.9\%$ for v2.

\begin{figure}[t]
\centering
\includegraphics[width=\columnwidth]{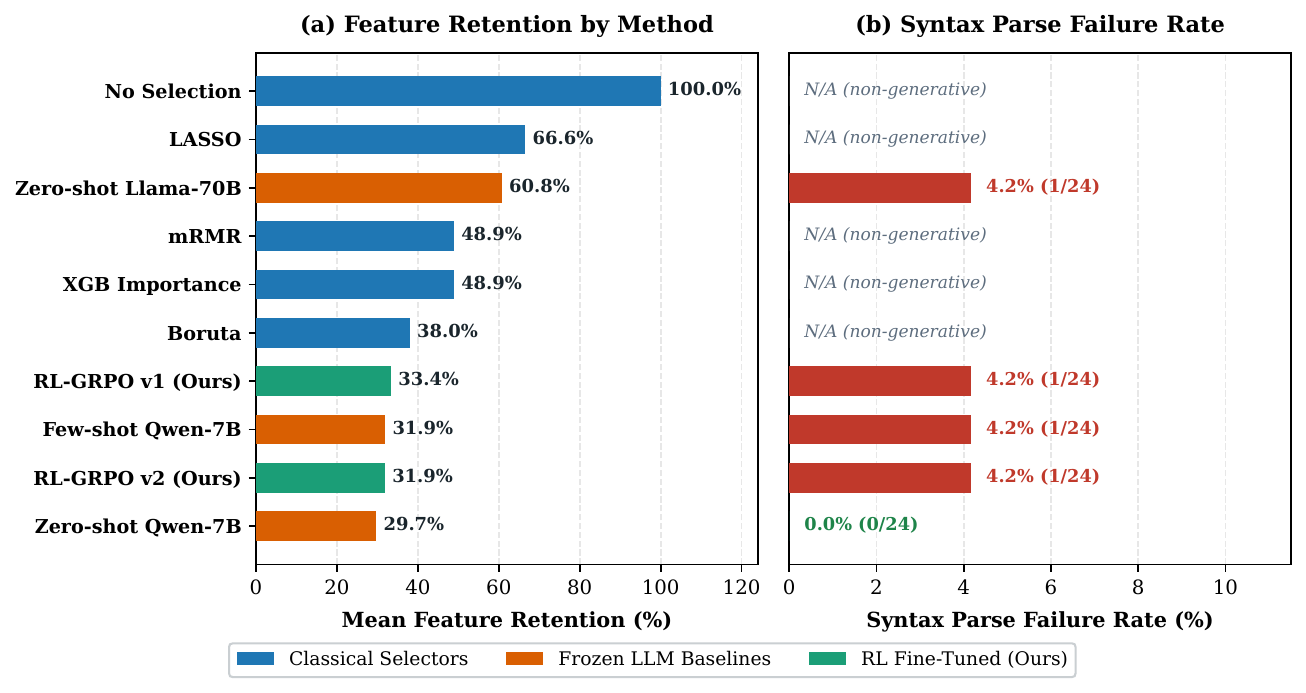}
\caption{Mean feature retention and syntax parse failure rates across the primary benchmark methods.}
\label{fig:retention}
\end{figure}

Figure~\ref{fig:retention} summarizes retention and parse reliability. The 7B selectors retain $29.7\%$--$33.4\%$ of features, versus $60.8\%$ for Llama-70B and $38.0\%$--$66.6\%$ for sparse classical selectors; parse failures remain rare across LLM/RL methods.

\begin{table}[t]
\centering
\caption{Matched-budget train-only classical control near RL-GRPO v1's $33.4\%$ retention budget over $N=5$ partitions (mean $\pm$ SD).}
\label{tab:matched_budget}
\setlength{\tabcolsep}{3.5pt}
\renewcommand{\arraystretch}{1.15}
\scriptsize
\begin{tabular}{l ccc}
\toprule
\textbf{Method} & \textbf{Mean Feat\%} & \textbf{Clf AUC (3)} & \textbf{Reg $R^2$ (3)} \\
\midrule
mRMR Top-$k$ (Train-Only) & $32.7\%$ & $0.872 \pm .016$ & $\mathbf{0.663 \pm .033}$ \\
XGB Imp.\ Top-$k$ (Train-Only) & $32.7\%$ & $\mathbf{0.890 \pm .017}$ & $0.646 \pm .028$ \\
LASSO Top-$k$ (Train-Only) & $32.7\%$ & $0.877 \pm .018$ & $0.631 \pm .034$ \\
\midrule
\textbf{RL-GRPO v1 (Ours)} & $33.4\%$ & $0.799 \pm .019$ & $0.596 \pm .015$ \\
\bottomrule
\end{tabular}
\end{table}

\textbf{Matched-Budget Train-Only Classical Control.}
Table~\ref{tab:matched_budget} shows that classical selectors remain stronger at approximately the same feature budget under strict train-only fitting. XGB Importance Top-$k$ achieves classification ROC-AUC $0.890\pm0.017$ versus $0.799\pm0.019$ for RL-GRPO v1, while mRMR Top-$k$ achieves regression $R^2$ $0.663\pm0.033$ versus $0.596\pm0.015$. The classical controls outperform v1 across all five partitions on five of six datasets; the exception is \textit{parkinsons\_telemonitoring}, where v1 reaches $0.888\pm0.012$ versus $0.859\pm0.011$ for XGB Top-$k$ and $0.855\pm0.036$ for mRMR Top-$k$. Thus, the classical advantage cannot be explained solely by larger subsets or full-dataset fitting.

\subsection{Effect of GRPO Training Configuration}
\label{sec:results:training}

Comparing RL-GRPO v1 and v2 shows a task-dependent effect. On classification, v2 improves on \textit{sonar} ($0.794$ vs.\ $0.726$) and slightly on \textit{adult\_income} ($0.856$ vs.\ $0.852$) but decreases on \textit{student\_dropout} ($0.801$ vs.\ $0.819$), raising mean ROC-AUC from $0.799$ to $0.817$. On regression, v2 performs worse on all three datasets---\textit{ames\_housing} ($0.860$ vs.\ $0.873$), \textit{parkinsons\_telemonitoring} ($0.883$ vs.\ $0.888$), and \textit{student\_performance} ($-0.043$ vs.\ $+0.027$)---lowering mean $R^2$ from $0.596$ to $0.567$ and worsening mean rank from $7.67$ to $9.17$. Thus, v2 provides no consistent held-out generalization improvement over validation-selected v1.

\subsection{Feature Semantics Analysis}
\label{sec:results:semantic}

\begin{figure*}[t]
\centering
\includegraphics[width=\textwidth]{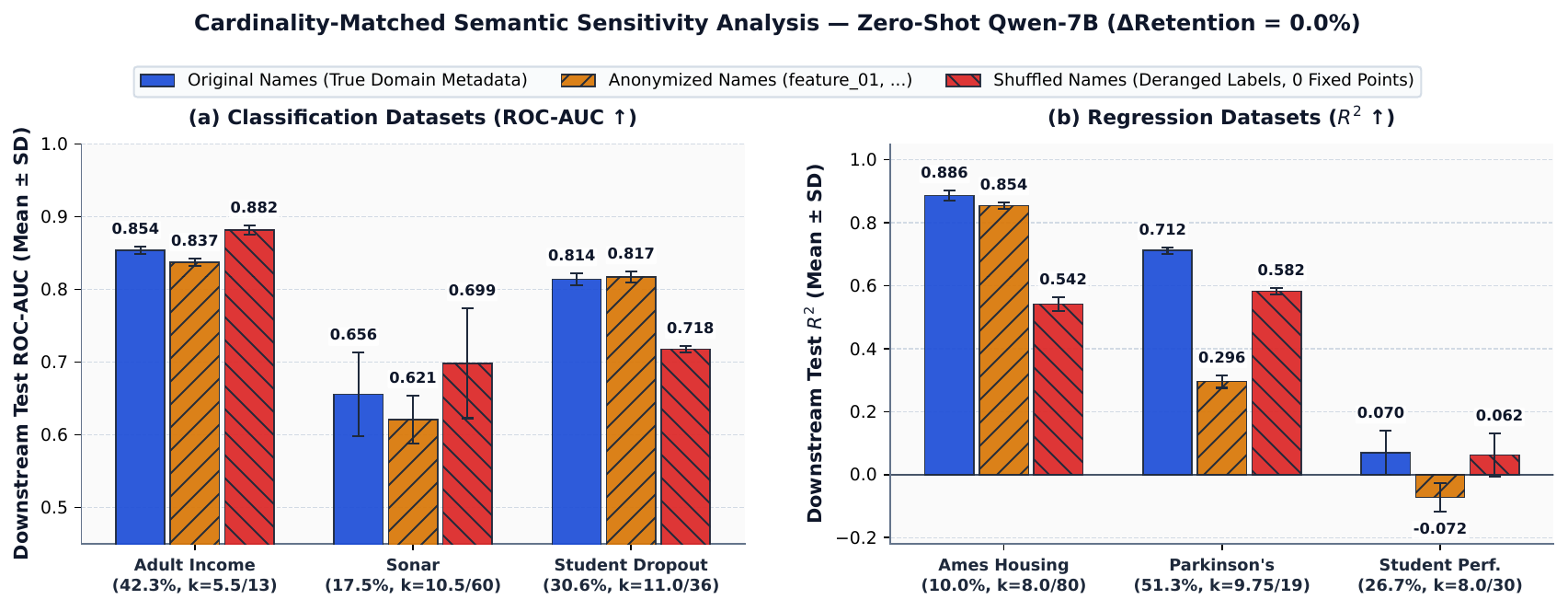}
\caption{Cardinality-matched semantic sensitivity of Zero-shot Qwen-7B across six held-out datasets over $N=5$ seeded evaluation partitions. Original, Anonymized, and Shuffled conditions use identical per-completion cardinality ($\Delta\mathrm{Retention}=0\%$).}
\label{fig:semantic}
\end{figure*}

Figure~\ref{fig:semantic} reports the cardinality-matched semantic diagnostic. Across regression datasets, Original feature names improve mean $R^2$ over Anonymized and Shuffled conditions by $+0.1964$ and $+0.1606$, respectively. Classification effects are smaller ($+0.0159$ and $+0.0084$ ROC-AUC) and dataset-dependent. Relative to Anonymized, Original increases mean regression $R^2$ from $0.359$ to $0.556$, with gains on \textit{parkinsons\_telemonitoring} ($+0.416$), \textit{student\_performance} ($+0.141$), and \textit{ames\_housing} ($+0.032$). Mean classification ROC-AUC increases more modestly from $0.759$ to $0.775$, while \textit{student\_dropout} remains near parity.

The strongest Original--Anonymized effect occurs on \textit{parkinsons\_telemonitoring}, where $R^2$ rises from $0.296$ to $0.712$ ($\Delta R^2=+0.416$). Shuffled names reveal sensitivity to misleading semantics: on \textit{ames\_housing}, performance falls from $0.886$ to $0.542$ ($\Delta R^2=-0.344$). Original and Shuffled selections have prompt-label Jaccard $0.363$ but true-column Jaccard $0.000$, indicating reliance on semantic labels even when detached from their columns. Overall, authentic semantics provide useful information at fixed cardinality, particularly for regression, but their effect remains dataset-dependent.

\section{Discussion}
\label{sec:discussion}

Overall, LLM-based feature selection is neither uniformly superior nor inferior to classical selection. Classical selectors generally achieve stronger predictive performance, including under a matched-sparsity, strictly train-only control. The cardinality-matched semantic analysis shows that feature-name semantics can materially influence LLM selections~\cite{choi2022lmpriors,jeong2024llmselect}, but their effect is dataset-dependent and does not close the performance gap with target-supervised statistical methods.

\subsection{When Do Feature Semantics Help?}
\label{sec:disc:works}

Semantic information is most useful when feature names carry meaningful domain information, particularly on regression tasks. The strongest effect occurs on \textit{parkinsons\_telemonitoring}, where authentic feature names yield substantially higher performance than anonymized names at identical cardinality. Smaller gains occur on several other datasets, while \textit{student\_dropout} shows near-parity. These results indicate that feature names can provide useful information beyond descriptive column statistics alone~\cite{hollmann2023caafe,jeong2024llmselect}.

However, semantic usefulness does not imply overall predictive superiority. Classical selectors retain stronger aggregate performance, including under the matched-budget train-only control, showing that useful semantic priors do not replace target-supervised statistical evidence.

\subsection{What Limits LLM Feature Selection?}
\label{sec:disc:fails}

The semantic and matched-budget controls highlight three mechanisms limiting metadata-driven LLM feature selection.

\textbf{1) Cardinality Confounding Can Mimic Semantic Failure:}
Changing feature names can also change how many features an LLM selects, confounding semantic effects with subset size. Exact-cardinality matching removes this source of variation, allowing differences to be attributed more directly to feature-name information.

\textbf{2) Mismatched Semantics Can Redirect Selection:}
When plausible feature names are deliberately assigned to the wrong columns, performance can degrade substantially. On \textit{ames\_housing}, Original and Shuffled selections retain prompt-label overlap (Jaccard $=0.363$) but no true-column overlap (Jaccard $=0.000$), showing that semantic labels can redirect selection even when detached from their underlying data.

\textbf{3) Absence of Target-Supervised Statistical Inspection:}
Prompt-based LLM selectors observe feature names and marginal summaries but not direct feature--target relationships or higher-order interactions. The matched-budget train-only control shows that classical selectors remain stronger at comparable sparsity, suggesting that access to target-supervised statistical information contributes to their advantage.

These findings suggest that feature semantics are best treated as complementary information rather than a substitute for target-supervised selection. Classical controls therefore remain important when evaluating LLM-based feature selectors, and alternative RL configurations should be selected using validation data before deployment.

\subsection{Limitations}
\label{sec:disc:limits}

We highlight five limitations:
\begin{enumerate}
  \item \textbf{Predictive Performance:} The GRPO policies do not outperform the strongest classical baselines overall; this study therefore focuses on characterizing when and how LLM-based selection behaves rather than claiming state-of-the-art predictive performance.
  \item \textbf{Single Base Architecture:} All RL fine-tuning experiments use Qwen2.5-7B-Instruct~\cite{qwen2025qwen25}.
  \item \textbf{Probe Model Scope:} The semantic diagnostic evaluates Zero-shot Qwen-7B; RL-adapted or larger models may exhibit different sensitivity profiles.
  \item \textbf{Evaluation Replication:} Evaluation uses $N=5$ seeded partitions, allowing variability estimates but providing limited power for formal paired significance testing; comparisons are therefore interpreted descriptively.
    \item \textbf{Feature-Dimensionality Scope:} The benchmark covers datasets with $\le 100$ features, so behavior in substantially higher-dimensional settings remains unresolved~\cite{guyon2003introduction,mcelfresh2023neural}.
\end{enumerate}
\section{Conclusion}
\label{sec:conclusion}

We presented a systematic study of LLM-based feature selection across an 18-dataset training/validation/held-out framework. Our conclusions align with our three contributions (Section~\ref{sec:intro}):

\begin{enumerate}
  \item \textbf{Controlled LLM feature-selection benchmark:} Across five seeded evaluation partitions, classical selectors generally achieve stronger predictive performance than the frozen and GRPO-adapted LLM selectors, while the 7B models produce compact subsets. This advantage persists under a strictly train-only matched-budget control.

  \item \textbf{Cardinality-matched semantic diagnostic:} At fixed subset cardinality, authentic feature names improve Zero-shot Qwen-7B selection most clearly on the regression tasks, while classification effects are smaller and dataset-dependent. Misleading feature names can also redirect selection toward poorer subsets.

\item \textbf{Robustness and matched-budget controls:} Task-separated metrics, cross-task ranks, and five seeded evaluation partitions show that GRPO adaptation produces task-dependent changes rather than consistent predictive gains. The v2 configuration does not consistently improve held-out performance over validation-selected v1.
\end{enumerate}

Overall, feature semantics provide useful side information but do not replace target-supervised statistical evidence. Future work should examine higher-dimensional datasets, richer semantic metadata, additional foundation models, and alternative RL adaptation strategies.
\bibliographystyle{IEEEtran}
\bibliography{references}

\end{document}